\documentclass[conference]{IEEEtran}
\IEEEoverridecommandlockouts
\usepackage[utf8]{inputenc}
\usepackage{amsmath, amssymb, amsfonts}
\usepackage{comment}
\usepackage{booktabs}
\usepackage{multirow}
\usepackage{tabularx}
\usepackage{stfloats}
\usepackage{array}
\usepackage{caption}
\usepackage{graphicx}
\usepackage{geometry}
\usepackage{setspace}
\usepackage{ragged2e}
\usepackage{booktabs}
\usepackage{pifont}
\usepackage{geometry}
\usepackage{array}
\newcommand{\cmark}{\ding{51}} 
\usepackage{tcolorbox}
\usepackage{tikz}

\usetikzlibrary{arrows, shapes, positioning, calc, mindmap,trees}
\usepackage{booktabs}
\usepackage{xcolor}
\definecolor{lightblue}{RGB}{240,248,255}
\usepackage{listings}
\usepackage{xcolor}
\usepackage{algorithmic}

\usepackage{hyperref}

\usepackage{cite}

\def\BibTeX{{\rm B\kern-.05em{\sc i\kern-.025em b}\kern-.08em
    T\kern-.1667em\lower.7ex\hbox{E}\kern-.125emX}}
\begin{document}

\title{Agentic AI Frameworks: Architectures, Protocols, and Design Challenges }

\author{\IEEEauthorblockN{Hana Derouiche}
\IEEEauthorblockA{\textit{University of Kairouan } \\
\textit{SMART Lab, University of Tunis}, Tunisia \\
hana.darouiche@gmail.com, 0009-0009-4162-5633}
\and
\IEEEauthorblockN{ Zaki Brahmi}
\IEEEauthorblockA{\textit{ University of Sousse} \\
\textit{Riadi Lab, Compus Manouba}, Tunisia \\
zakibrahmi@gmail.com, 0000-0002-0432-4817}
\and
\IEEEauthorblockN{ Haithem Mazeni }
\IEEEauthorblockA{\textit{ University of Jandouba}, Tunisia \\
haithem.mezni@gmail.com, 0000-0001-9932-8433}

}

\maketitle
\begin{abstract}
The emergence of Large Language Models (LLMs) has ushered in a transformative paradigm in artificial intelligence, Agentic AI, where intelligent agents exhibit goal-directed autonomy, contextual reasoning, and dynamic multi-agent coordination. This paper provides a systematic review and comparative analysis of leading Agentic AI frameworks, including CrewAI, LangGraph, AutoGen, Semantic Kernel, Agno, Google ADK, and MetaGPT, evaluating their architectural principles, communication mechanisms, memory management, safety guardrails, and alignment with service-oriented computing paradigms.
Furthermore, we identify key limitations, emerging trends, and open challenges in the field. To address the issue of agent communication, we conduct an in-depth analysis of protocols such as the Contract Net Protocol (CNP), Agent-to-Agent (A2A), Agent Network Protocol (ANP), and Agora.
Our findings not only establish a foundational taxonomy for Agentic AI systems but also propose future research directions to enhance scalability, robustness, and interoperability. This work serves as a comprehensive reference for researchers and practitioners working to advance the next generation of autonomous AI systems.
\end{abstract}

\begin{IEEEkeywords}
Agentic AI, Large Language Models, Agent protocols, Agentic AI-as-a-Service
\end{IEEEkeywords}

\section{Introduction}

\footnote{© 2025 IEEE. Personal use of this material is permitted. Permission from IEEE must be obtained for all other uses, in any current or future media, including reprinting/republishing this material for advertising or promotional purposes, creating new collective works, for resale or redistribution to servers or lists, or reuse of any copyrighted component of this work in other works}The rapid advancement of Large Language Models (LLMs) has ushered in a new era of intelligent agents, known as Agentic AI, where autonomous systems, referred to as intelligent agents, can reason, communicate, and coordinate to complete complex, long-horizon tasks. This paradigm shift departs from traditional AI and Multi-Agent Systems (MAS) \cite{ferber1999multi} by introducing agents that are not only context-aware but also capable of goal-directed behavior powered by LLM-based cognition.

Agentic AI is increasingly being deployed in domains such as software engineering \cite{bornet2025agentic}, scientific discovery, business automation, and human-agent collaboration. To support its capabilities, a growing ecosystem of Agentic AI frameworks has emerged (e.g., CrewAI, LangGraph). These frameworks provide architectural foundations and tooling for building, orchestrating, and deploying intelligent agents.
Despite the rapid growth of the Agentic AI paradigm, there remains a lack of systematic understanding of how these frameworks differ in their design philosophies, technical components, and practical capabilities. To our knowledge, the existing literature on this topic remains scarce and often focuses on isolated features. For instance, authors in \cite{joshi2025advancing} provide a comprehensive review in the context of financial services.  

This paper aims to bridge the gap by offering a comprehensive comparative analysis of leading frameworks such as CrewAI, LangGraph, AutoGen, Semantic Kernel, and MetaGPT. Our study is based on an exploration of the architectural features that characterize major Agentic AI frameworks, highlighting their design patterns and operational components. Attention is also given to the communication protocols (e.g., ACP, ANP, A2A, Agora) adopted by these systems.
In addition, the paper investigates how different frameworks handle critical aspects such as memory integration and guardrail enforcement. Finally, it reflects on the current limitations and challenges these systems face, while identifying promising directions for future development in Agentic AI. To this end, we address the following research questions:

\begin{itemize}
    \item \textbf{RQ1:} How have intelligent agents evolved from traditional AI agents to modern LLM-powered agents?
  \item \textbf{RQ2:} What frameworks are available for developing agentic AI systems, and how do they implement core agent concept, MAS paradigms (negotiation, collaboration, organization), and communication?
  \item \textbf{RQ3:} How do these frameworks compare in communication, memory, orchestration, modularity, and guardrails? What recent advances exist in agent communication protocols?
   \item \textbf{RQ4:} To what extent are modern agentic AI frameworks ready for integration into service computing ecosystems?
\end{itemize}

The remainder of the paper is organized as follows: Section II discusses the foundations of intelligent agents and communication protocols. Section III examines communication protocols in greater detail. Section IV analyzes Agentic AI frameworks with respect to memory, guardrails, and service computing. Section V outlines current limitations and open research directions. Section VI concludes the paper.

\section{Intelligent Agent}

The concept of an "agent" in artificial intelligence has evolved significantly over the past decades within foundational paradigms of AI, primarily Multi-Agent Systems (MAS) and expert systems \cite{ren2004multi}. Traditionally, an agent was defined as an autonomous entity capable of perceiving its environment through sensors and acting upon it through effectors to achieve designated goals. This classical definition emphasized autonomy, reactivity, proactivity, and social ability, core principles in early MAS research \cite{ferber1999multi}.
However, with the rise of Large Language Models (LLMs) and transformer-based architectures, modern agents exhibit more dynamic and context-aware behaviors. They are no longer confined to predefined environments but instead operate within fluid, often human-centered contexts. These agents not only reason and act but also interact with external data sources, orchestrate tools, and collaborate with other agents in real time, often asynchronously.

Contemporary agent architectures, including ReAct \cite{yao2022react}, PRACT \cite{liu2024pract}, RAISE \cite{raise2024}, and Reflexion \cite{shinn2023reflexion}, are unified by their reliance on LLMs as reasoning engines, orchestrating planning, memory, dialogue, and tool use through iterative loops. For instance, the ReAct architecture combines Reasoning (chain-of-thought) and Acting (tool use) in an iterative loop.

To break it down, we believe that \textit{modern agents fundamentally differ from classical agents (e.g., Belief-Desire-Intention (BDI) agents) by leveraging LLMs and advanced technologies as versatile reasoning engines and dynamic tool portfolios.} Table \ref{tab:agent_comparison} presents a comparison between traditional and modern AI agents.

\begin{table*}[!t]
\centering
{\scriptsize
\caption{Traditional AI agents vs. Modern AI agents}
\label{tab:agent_comparison}
\begin{tabular}{p{1.75cm}p{6.5cm}p{7cm}}
\hline
\textbf{Aspect} & \textbf{Traditional AI agents} & \textbf{Modern agentic AI systems (LLM-based agents)} \\
\hline
Definition & Autonomous entities with fixed sensing/acting loops; limited by static rules or models & Autonomous reasoning systems using LLMs with dynamic behavior, tool orchestration, and context-awareness \\
\hline
Autonomy & Limited autonomy; often dependent on human input or predefined instructions & High autonomy; capable of independently performing complex and extended tasks \\
\hline
Goal Management & Focused on single, static goals or fixed task planning & Capable of managing multiple, evolving, and nested goals adaptively \\
\hline
Architecture & Rule-based or BDI (Belief–Desire–Intention) models; monolithic design & Modular architecture centered on LLMs, with components for memory, tools, context injection, and roles \\
\hline
Adaptability & Suited to controlled, predictable environments; poor generalization & Designed for open, dynamic, and unpredictable environments \\
\hline
Decision-Making & Deterministic or rule-based logic; symbolic reasoning & Context-sensitive, probabilistic reasoning with adaptive planning and self-reflection \\
\hline
Learning Mechanism & Rule-based or supervised learning with limited updates & Self-supervised and reinforcement learning; continual fine-tuning possible \\
\hline
Context Handling & Static or manually coded states and rules & Dynamic context injection via agent protocols (e.g., MCP, A2A) and runtime awareness \\
\hline
Communication & Message-passing via ACL or KQML & Real-time, event-driven collaboration; natural language interfaces \\
\hline
Tool Use & Limited or predefined tools and actions & Dynamic tool invocation, chaining, and API calling based on context \\
\hline
Memory & Optional, often hardcoded or task-specific & Integrated memory systems supporting long- and short-term information retention \\
\hline
\end{tabular}
}
\end{table*}

Given this broad evolution, it is now necessary to rethink and potentially redefine what constitutes an agent. A modern agent may be better defined as: \textit{"An autonomous and collaborative entity, equipped with reasoning and communication capabilities, capable of dynamically interpreting structured contexts, orchestrating tools, and adapting behavior through memory and interaction across distributed systems."}


\section{Agent Communication Protocols}

The rise of LLM-powered autonomous agents has highlighted critical challenges in interoperability, security, and scalability, largely due to fragmented frameworks and ad hoc integrations \cite{wang2024survey,yang2025survey}. Robust agent communication protocols are essential for enabling peer discovery, context sharing, and coordinated action, forming the backbone of modular and resilient Multi-Agent Systems. These protocols offer clear advantages over traditional interaction models.
Agent communication protocols have evolved from early semantic standards such as FIPA ACL in the 1980s–1990s, to web-based systems (e.g., SOAP/WSDL) in the 2000s–2010s, culminating in today's LLM-driven protocols (e.g., ACP, ANP) and prospective neuro-symbolic or quantum-secure architectures.
Despite their transformative potential, clear and universally adopted standards remain nascent, creating a gap that hinders the scalability and composability of multi-agent ecosystems \cite{yadav2020distributed,ray2025survey}. Emerging protocols (e.g., MCP, A2A, Agora) aim to bridge this gap through lightweight JSON-RPC schemas for context exchange, performative messaging, and discovery.

Fundamentally, contemporary communication protocols share a unifying principle: ``\textit{eliminate the need for manual integration, custom middleware, or deep protocol-specific expertise by providing standardized, intelligent frameworks for seamless interaction between agents, whether in AI-to-AI, agent-to-network, or multi-agent systems}.''
One of the earliest protocols, the \textbf{Model Context Protocol (MCP)}\footnote{\url{https://modelcontextprotocol.io/introduction}, accessed 10-05-2025}, was initially designed for structured tool calls via JSON-RPC and secure schema validation. Although MCP follows a client–server model, it can support inter-agent delegation where strict hierarchical roles are required.
Later, \textbf{Google’s Agent2Agent Protocol (A2A)} \cite{google2025a2a} introduced a more agent-oriented architecture, enabling capabilities such as memory management, goal coordination, task invocation, and capability discovery. A2A formalizes communication through constructs like Agent Cards, Task Objects, and Artifacts (standardized outputs).
To support decentralized identity and semantic interoperability, the \textbf{Agent Network Protocol (ANP)} \cite{anp2024} incorporates decentralized identifiers (DIDs) and JSON-LD semantics, organizing communication around a lifecycle (creation, operation, update, termination) \cite{ehtesham2025survey}. It accommodates both explicitly defined protocols and natural language negotiation using LLMs. 
Built on similar principles, the \textbf{Agent Communication Protocol (ACP)}\footnote{\url{https://agentcommunicationprotocol.dev/}}, originally started at IBM, allows agents to communicate via RESTful APIs, using structured JSON messages to encode actions, goals, and intents. Its design is transport-agnostic and compatible with Web3 environments, making it suitable for scalable, cross-organizational communication.
At a higher level of abstraction, \textbf{Agora}\footnote{\url{https://agoraprotocol.org/}, accessed 10-05-2025} \cite{marro2024scalable} serves as a meta-coordination layer, integrating multiple protocols including MCP, ANP, and ACP. It introduces Protocol Documents (PDs), which are machine-interpretable specifications that guide agents in selecting or constructing communication protocols.
Table \ref{tab:Agentic_AI_Protocols} presents a comprehensive comparison of the studied protocols based on criteria including discovery, messaging, layering, etc.

\begin{table*}[t]
{\scriptsize
\caption{Comparison of modern agentic AI protocols}
\label{tab:Agentic_AI_Protocols}
\resizebox{\textwidth}{!}{ 
\begin{tabular}{>{\raggedright\arraybackslash}p{2cm}>{\raggedright\arraybackslash}p{2.25cm}>{\raggedright\arraybackslash}p{3.4cm}>{\raggedright\arraybackslash}p{3cm}>{\raggedright\arraybackslash}p{3cm}>{\raggedright\arraybackslash}p{2.75cm}}
\toprule
\textbf{Feature} & \textbf{MCP} & \textbf{ACP} & \textbf{A2A} & \textbf{ANP} & \textbf{Agora} \\
\midrule

\textbf{Message Format} & 
JSON-RPC & 
JSON-LD & 
JSON-RPC/HTTP/SSE & 
JSON-LD + NLP & 
PD + Natural Language \\ \hline

\textbf{Semantics} & 
Custom performatives & 
Goal-oriented messages (e.g., goal, action) & 
Custom performatives & 
PD & 
PD \\ \hline

\textbf{Discovery} & Manual & 
Agent metadata (agent.yml) and Registry  & 
Agent Card & 
Agent description as JSON-LD & 
Exchanging natural-language PDs \\ \hline

\textbf{Frameworks} & 
LangChain, OpenAgents, Agno & 
AutoGen, LangGraph, CrewAI & 
AutoGen, CrewAI, LangGraph & 
AGORA, CrewAI, Semantic Kernel Agent & 
-  \\ \hline
\textbf{Transport Layer} & 
HTTP, Stdio, SSE & 
HTTP & 
HTTP, optional SSE & 
HTTP with JSON-LD & 
HTTP with PD \\ \hline

\textbf{Use Case} & 
LLM-tool integration & 
Cross-agent collaboration & 
Enterprise agent orchestration & 
Decentralized agent markets & 
Multi-agent environments \\

\bottomrule
\end{tabular}
}}
\end{table*}

\begin{tcolorbox}[
    colback=gray!10,
    colframe=black,
    title=Key Findings,
    sharp corners,
    boxrule=0.5pt,
    fonttitle=\bfseries
]
\footnotesize

Modern agentic protocols (MCP, ACP, A2A, ANP, Agora) reflect a shift toward service-oriented interoperability, with JSON-LD/PD semantics enabling dynamic discovery and composition. Yet, fragmentation persists, HTTP dominates transport, but semantic heterogeneity (custom performatives versus goal-oriented/PD messages) limits seamless integration. Frameworks like AutoGen bridge domains, but standardized service contracts (akin to WSDL for agents) remain nascent, hindering large-scale agent-as-a-service adoption.

\end{tcolorbox}

\section{Agentic AI Frameworks}

\subsection{Comparative overview}

\begin{figure}[!t]
    \centering
    \vspace{-1em}  
    \includegraphics[width=\linewidth]{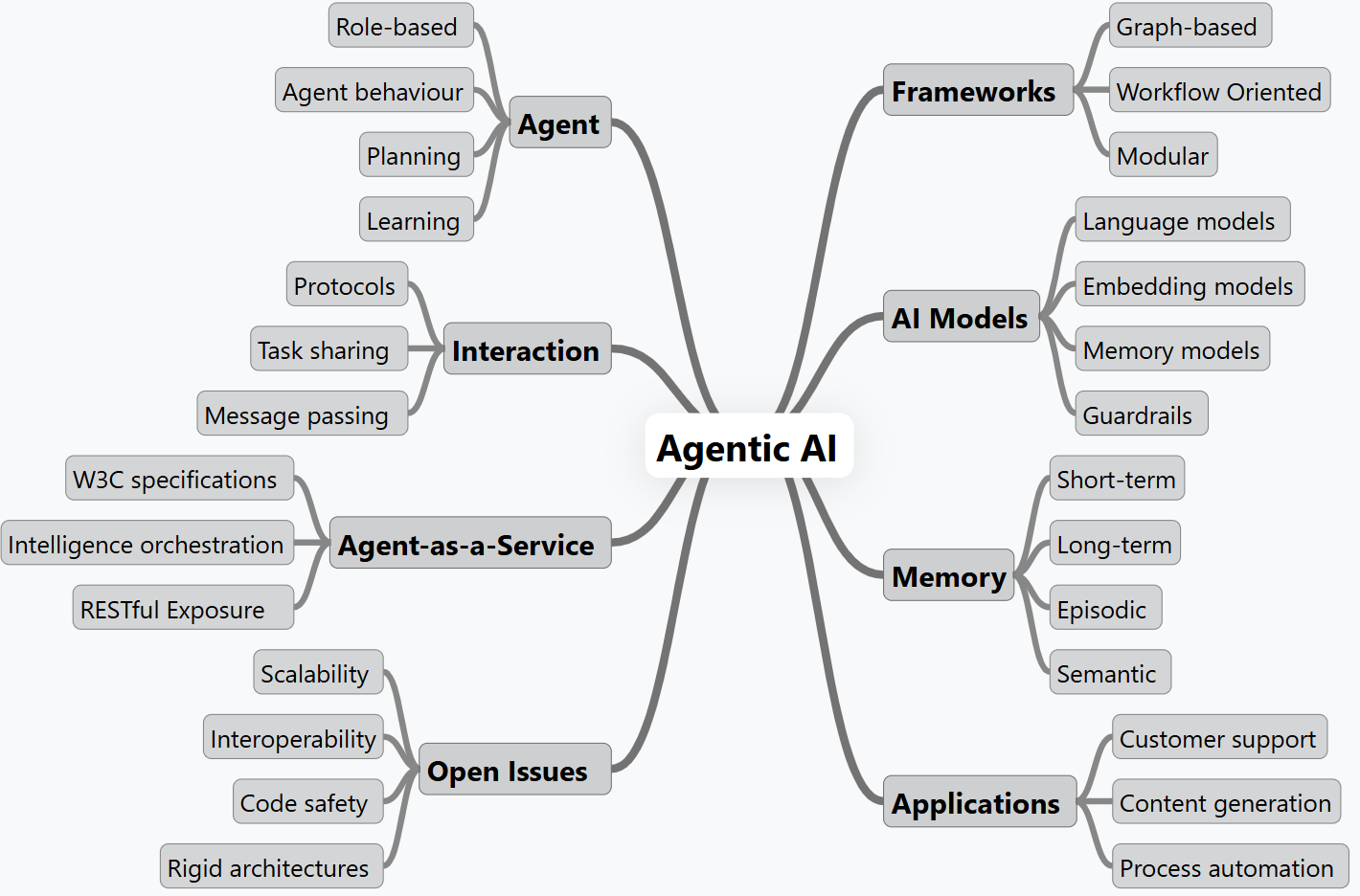}
    \caption{\small Agentic AI design taxonomy}
    \label{fig:taxonomy}
    \vspace{-1em}  
\end{figure}

Agentic AI frameworks provide foundational infrastructure for developing systems where agents exhibit autonomy, context-awareness, and goal-directed behavior. These agents, powered by LLMs, dynamically interpret tasks, orchestrate tool use, and adapt to real-time environments. In this section, we synthesize major agentic frameworks by classifying them based on shared principles and usage patterns, highlighting how their design choices shape agent behavior and coordination (see Fig. \ref{fig:taxonomy}).

Several frameworks focus on structured orchestration and multi-agent workflows. \textbf{AutoGen} \cite{wu2023autogen}, developed by Microsoft, enables rich multi-agent conversations with shared tools and modular LLM backends. It provides the backbone for collaborative workflows across domains such as coding and automation. Similarly, \textbf{CrewAI} \cite{duan2024exploration} promotes role-based collaboration among agents, emphasizing coordination and delegation for team-based problem-solving. The listing \ref{lst:crew_agent} shows an example of \verb|crewAI| agent. 

\lstset{
  language=Python,
  basicstyle=\ttfamily\small,
  keywordstyle=\color{blue},
  commentstyle=\color{gray},
  stringstyle=\color{orange},
  frame=single,
  breaklines=true,
  columns=flexible
}

\begin{lstlisting}[basicstyle=\footnotesize\ttfamily, caption={Simple CrewAI Agent}, label={lst:crew_agent}]
agent = Agent(
     role="Research Assistant",
     goal="Summarize recent AI news",
     backstory="An AI expert who keeps track of the latest in research.",
     llm=OpenAI(temperature=0.5),
     tools=[],
     memory=True
 )
\end{lstlisting}

Another framework, \textbf{MetaGPT} \cite{hong2023metagpt}, follows a comparable philosophy by simulating real-world software engineering teams, where each agent adopts a specialized role (e.g., project manager or developer) to perform structured tasks in a product lifecycle pipeline.
For lightweight and transparent agent composition, \textbf{SmolAgents} and \textbf{PydanticAI}\footnote{\url{https://ai.pydantic.dev/}, accessed 10-05-2025} provide minimal yet effective solutions. SmolAgents emphasizes simplicity and modularity, supporting prompt chaining and tool use with low overhead. PydanticAI uses the Pydantic library to define agent schemas, enhancing reproducibility and safety, especially for debugging and deployment.

In terms of orchestration abstraction and development ease, the \textbf{OpenAI Agents SDK} provides a high-level interface that encapsulates tool use, memory, and instruction-following behavior. Other frameworks lean toward graph-based or declarative orchestration. \textbf{LangGraph} \cite{wang2024agent} introduces a novel graph-based model for sequencing tasks among LLM agents. By supporting compositional flows and stateful operations, it allows for traceable and scalable agent design, particularly in research and analytics contexts. Along similar lines, \textbf{Semantic Kernel} \cite{soh2024semantic} provides enterprise-grade orchestration with fine-grained control over planning, memory, and skill execution, enabling integration with external systems in structured reasoning scenarios. \textbf{Agno}, meanwhile, promotes a declarative and transparent approach to defining agent goals, tools, and reasoning logic, making it a strong candidate for automation workflows requiring explainability and control.

Finally, frameworks like \textbf{LlamaIndex} and \textbf{Google ADK} push the boundaries of data-centric and distributed agent ecosystems. \textbf{LlamaIndex} empowers agents with capabilities for querying structured and unstructured data for knowledge-intensive applications. \textbf{Google ADK}, still experimental and designed for scalability, allows orchestration of multi-agent workflows, making it suitable for adaptive AI assistants and enterprise automation.

\begin{figure}[ht]
    
    \includegraphics[width=0.6\textwidth]{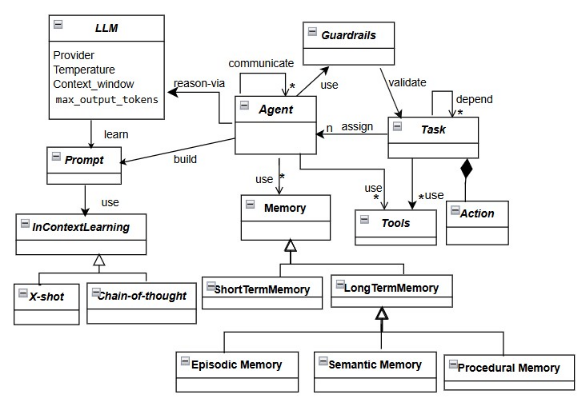}
    \caption{Unified class model for Agentic AI frameworks}
    \label{fig:agent_class_diagram}
\end{figure}

To distill a generic and reusable agent model by identifying common structural patterns, the proposed class diagram in Fig.~\ref{fig:agent_class_diagram} schematizes a unified class model.

\begin{tcolorbox}[
    colback=gray!10,
    colframe=black,
    title=Key Findings,
    sharp corners,
    boxrule=0.5pt,
    fonttitle=\bfseries
]
\footnotesize
In practice, frameworks share core components. The \textit{LLM} enables advanced reasoning through prompt-based interactions enhanced by in-context learning (few-shot, one-shot, chain-of-thought prompting), allowing agents to perform complex cognitive tasks with minimal supervision; \textit{tools} (external actions); \textit{memory}; and \textit{guardrails} to ensure safety, reliability, and validation of agent outputs and actions.

\end{tcolorbox}

\subsection{Memory in Agentic AI frameworks}

Memory is foundational to agentic AI, enabling context-aware, adaptive behavior \cite{guo2023empowering}. Its mechanisms support retention, retrieval, and reasoning across interactions, facilitating multi-turn dialogues, preference adaptation, and knowledge transfer. Memory can be mainly categorized into (1) \textit{short-term memory}, which allows agents to maintain the immediate conversational or task context, and (2) \textit{long-term memory}, which, by contrast, captures persistent data across sessions, such as user preferences, task history, or learned knowledge, that agents can revisit later. Some frameworks also implement specialized forms of long-term memory, such as \textit{semantic memory} \cite{sarthou2019ontologenius}, which stores and reuses past reasoning paths or decisions; \textit{procedural memory}, which recalls specific task flows or strategies previously used; and \textit{episodic memory} \cite{dechant2025episodic}, which encodes detailed contextual snapshots of specific past interactions or experiences, enabling more nuanced and personalized agent behavior over time \cite{nuxoll2012enhancing}.

Across the surveyed frameworks, memory is implemented in various ways depending on the target use case and design philosophy. For instance, LangGraph integrates memory as part of its graph-based structure, preserving state within and across nodes, thereby enabling agents to follow structured workflows with context retention. OpenAI’s SDK supports memory through conversation sessions, maintaining task-specific state implicitly, which simplifies implementation for developers. CrewAI equips the agent with individual memory, which plays a central role in role-specific coordination and delegation. AutoGen supports structured dialogues among agents where memory can be passed, persisted, or modified across roles \cite{wu2023autogen}. Google ADK maintains shared memory for dynamic collaboration and task handovers. In contrast, Agno employs a more declarative memory approach to enhance transparency and inspectability.

Table \ref{tab:memory_comparison} provides a comparative overview of memory support across these frameworks, based on their official documentation and observed implementation patterns.

\begin{table*}[ht]
{\scriptsize
\centering
\caption{Memory support in Agentic AI frameworks}
\label{tab:memory_comparison}
\begin{tabular}{p{1.7cm}p{6.5cm}ccccc}
\hline
\textbf{Framework} & \textbf{Memory Approach} & \textbf{Short-Term} & \textbf{Long-Term} & \textbf{Semantic} & \textbf{Procedural} & \textbf{Episodic} \\
\hline
LangGraph & Stateful graph nodes retain context between agent transitions. & \checkmark & -- & -- & -- & -- \\
\hline
OpenAI SDK & Session-based memory abstraction (e.g., \texttt{ConversationBufferMemory}). & \checkmark & -- & -- & -- & -- \\
\hline
SmolAgents & memory is optional and manually injected. & -- & -- & -- & -- & -- \\
\hline
CrewAI & Agent-level memory for dialogue and coordination, with entity/contextual memory. & \checkmark & \checkmark & \checkmark & -- & \checkmark \\
\hline
AutoGen & Shared memory context maintained across structured dialogues. & \checkmark & \checkmark & -- & -- & \checkmark \\
\hline
Semantic Kernel & Extensible memory modules integrated with planners and skills. & \checkmark & \checkmark & \checkmark & \checkmark & -- \\
\hline
LlamaIndex & Embedding-based context retrieval from large-scale indexed data. & \checkmark & \checkmark & \checkmark & -- & -- \\
\hline
PydanticAI & Schema-first modeling; external memory systems can be attached. & -- & -- & -- & -- & -- \\
\hline
Google ADK & Shared memory across agent instances and system modules. & \checkmark & \checkmark & -- & -- & -- \\
\hline
Agno & Declarative memory structure embedded in agent design. & \checkmark & -- & -- & -- & -- \\
\hline
MetaGPT & Implicit memory through role-based behavioral. & \checkmark & \checkmark & \checkmark & \checkmark & -- \\
\hline
\end{tabular}
}
\end{table*}

\subsection{Guardrails in Agentic AI Frameworks}

Guardrails ensure AI agents act safely and predictably by validating outputs, enforcing security, and maintaining workflow integrity. Among current frameworks, AutoGen, LangGraph, Agno, and the OpenAI SDK provide the strongest native support. AutoGen includes validators and retry logic; LangGraph enables advanced flow-level checks via node validation; Agno offers an early-stage trust layer; and the OpenAI SDK supports schema validation with developer-defined safeguards. Others like CrewAI, MetaGPT, and Google ADK provide partial support, while LlamaIndex and Semantic Kernel validate only at specific stages. SmolAgents lacks guardrails entirely, prioritizing developer control over safety.
Overall, while guardrail capabilities are emerging, most frameworks require external logic or manual setup for robust enforcement. This highlights a need for standardized, modular safety layers in agentic AI development.

\subsection{Applications of Agentic AI frameworks}

Agentic AI frameworks like CrewAI and LangGraph have been applied across domains to coordinate specialized LLM agents. In finance, they support tasks such as risk management, anomaly detection, and strategy development through multi-agent collaboration \cite{joshi2025comprehensive,okpala2025agentic}. CrewAI enables reasoning over historical data for informed decision-making. LangGraph has been used in intelligent transportation for modular traffic management \cite{chen2025implementing}, while CrewAI also supports automated travel planning in tourism by enabling agents to analyze cities and plan itineraries collaboratively \cite{singh2024automated}.

Despite these efforts, broader adoption of agentic AI frameworks faces challenges. Key barriers include a lack of architectural transparency and standardization, as most solutions lack reusable, interoperable designs like those found in service-oriented systems. Leading frameworks (e.g., AutoGen, AutoGPT) remain underutilized in domain-specific fields (e.g., finance, healthcare). Additionally, multi-agent coordination protocols are often inadequate, scalability is limited, and standardized APIs for collaboration are urgently needed (see Section~\ref{sec:agenticai-soc}).

\subsection{Agentic AI from a service computing perspective}
\label{sec:agenticai-soc}

This section addresses RQ4: \textit{To what extent are agentic AI frameworks ready for integration into service-computing ecosystems?} We evaluate their maturity in Table~\ref{tab:agentic_service} by analyzing key service-oriented capabilities, such as dynamic discovery, composition, and orchestration, against the requirements of modern service architectures.

\begin{table*}[ht]
{\scriptsize
\centering
\caption{Compatibility of Agentic AI frameworks with core service computing functions}
\label{tab:agentic_service}
\begin{tabular}{p{1.5cm}p{1.3cm}p{1.4cm}p{1.3cm}p{10cm}}
\hline
\textbf{Framework} & \textbf{Discovery} & \textbf{Publishing} & \textbf{Composition} & \textbf{Key Observations} \\
\hline
CrewAI     & $\times$ & $\times$ & \checkmark & Role-based agents with task delegation; requires external registry for discovery and publishing. \\
\hline
LangGraph       & \checkmark\textsuperscript{a} & $\times$ & \checkmark & State-machine logic allows robust composition; discovery possible via extension hooks. \\
\hline
AutoGen         & $\times$ &$\times$ & $\sim$ & Conversational agents can invoke tools sequentially; limited planning logic. \\
\hline
Semantic Kernel & Partial\textsuperscript{a} & Partial\textsuperscript{b} & \checkmark & Supports dynamic composition via planners, but discovery and publishing mechanisms require external implementation or integration. \\

\hline
Agno     & $\times$ & $\times$  & $\times$ & Minimalist reasoning layer; requires external logic for composition. \\
\hline
Google ADK      & Partial\textsuperscript{a} & Partial\textsuperscript{a} & \checkmark & Service discovery and publishing require integration with Google Cloud services such as API Gateway and Service Directory. \\
\hline
MetaGPT         & $\times$ & $\times$ & $\sim$ & Generates orchestrators and workflows in code; lacks runtime execution support. \\
\hline
\end{tabular}

\footnotesize\textsuperscript{a}\,Achieved by connecting graph transitions to an external service catalog. }
\end{table*}

Semantic Kernel and Google ADK offer strong support for service composition through skill planners and cloud integration, respectively. However, neither framework embeds full service computing primitives natively. Their readiness depends on integration with external registries and orchestration layers. LangGraph, with its state machine abstraction, also provides robust composition patterns and extensibility hooks for discovery. LangGraph offers deterministic, fault-tolerant orchestration and can support discovery through simple catalog adapters, making it a strong runner-up. 
By contrast, CrewAI, AutoGen, Agno, and MetaGPT excel at multi-agent planning or code generation but require an auxiliary service registry (e.g., OpenAPI gateway or service mesh) to participate in fully dynamic service ecosystems. Incorporating such registries would elevate these frameworks from task-centric agent platforms to comprehensive service-computing solutions.

To support \textit{service-oriented Agentic AI}, current frameworks have begun integrating W3C standards (e.g., WSDL, WS-Policy, BPEL), but adoption remains limited (see Table \ref{tab:w3c-agenticai}). JSON-schema function registration in CrewAI and the OpenAI SDK mimics WSDL, and AutoGen reflects BPEL-style orchestration without declarative syntax. WS-Policy and WS-Security principles appear in Agno and SmolAgents via runtime settings and JWTs, though they lack formal policy or security token formats. Coordination logic and SLA-like behavior exist in frameworks like MetaGPT and CrewAI, yet without formal constructs for WS-Coordination or WS-Agreement. Overall, W3C-inspired features are emerging, but standardized, interoperable adoption is still lacking.

\begin{table*}[!t]
\scriptsize
\caption{W3C specifications and their adaptation for Agentic AI frameworks}
\label{tab:w3c-agenticai}
\centering
\begin{tabular}{p{1.1cm} p{2.3cm} p{4cm} p{4.5cm} p{4cm}}
\hline
\textbf{Spec.} & \textbf{Role in Agentic AI} & \textbf{Integration benefits} & \textbf{Managed AI entities} & \textbf{Current support} \\
\hline

WSDL &
Describes agent function contracts &
Enables discoverability of agent capabilities and explicit API documentation &
\texttt{<portType>} describes an agent/tool endpoint; \texttt{<operation>} names a callable function; \texttt{<binding>} maps to API or model invocation. &
\textit{CrewAI} and \textit{OpenAI SDK} are limited to JSON schema for functions wrapping  and registration. \\ \hline

BPEL &
Orchestrates multi-agent workflows &
Enables structured planning and execution of agent-based tasks, error handling, and workflow modularity &
\texttt{<process>}, \texttt{<sequence>}, \texttt{<invoke>} reflect agent invocation sequences and transitions (planner/executor/critic roles). &
Multi-agent workflows in \textit{AutoGen} \cite{wu2023autogen}. \\ \hline

WS-Policy &
Controls agent runtime configurations. &
Enforcement of runtime constraints across agents and tasks, allowing dynamic configurability. &
\texttt{<Policy>}, \texttt{<All>}, \texttt{<ExactlyOne>} model parameter sets (e.g., temperature, max tokens) of agent tools and behaviors. &
Per-agent runtime policy integration in \textit{Agno}, Per-call parameter control in \textit{OpenAI SDK}. \\ \hline

WS-Security &
Secures inter-agent communications and authenticates actions. &
Ensures confidentiality of exchanged prompts, provenance of agent-generated content, integrity of inter-agent communication. &
\texttt{<SecurityToken>}, \texttt{<EncryptedData>}, \texttt{<Signature>} protect agent messages and signed prompts. &
JWTs- and encryption- based in inter-agent messaging in \textit{SMOLAgent}. \\ \hline

WS-Coordination &
Manages session context, turn-taking, and agent roles &
Coordination of agent sessions, including turn-taking, role enforcement, and shared context propagation. &
\texttt{<CoordinationContext>}, \texttt{<Register>} track sessions and dialog flow between named agents. &
Agent SOPs with distinct roles in MetaGPT, 
Agent in CrewAI are defined by role and turn policies \\ \hline

WS-Agreement &
Negotiates QoS among agents &
Supports performance-aware selection and delegation of agents, by expressing SLA guarantees. &
\texttt{<ServiceDescriptionTerm>}, \texttt{<GuaranteeTerm>} express agent expectations and SLAs for selection. &
AutoGen planner selects agents by estimated criteria, CrewAI priorities influence selection. \\ \hline

\hline
\end{tabular}

\end{table*}
\section{Limitations and Challenges}
Despite rapid progress, current agentic AI frameworks exhibit several critical limitations. These limitations  span architectural rigidity, dynamic collaboration constraints, safety risks, and lack of interoperability.

\textbf{Rigid architectures}: Most frameworks enforce static agent roles (e.g., planner, executor, coder), which limits adaptability in dynamic or evolving tasks. For instance, in MetaGPT or CrewAI, once an agent is assigned a predefined role, it cannot easily change behavior during execution.

\textbf{No runtime discovery}: Agents in many systems cannot dynamically discover or collaborate with peers during runtime. Instead, all agent interactions must be statically defined, limiting scalability and emergent cooperation. As a solution, we can implement an \textit{agent or skill registry}, a central directory where agents can publish and query capabilities. This allows new agents to join the system and form collaborations dynamically.

\textbf{Code safety}: Execution of generated code, which is common in MetaGPT and AutoGen, poses severe safety risks. Generated Python code can include file system access, shell commands, or unsafe imports. To ensure secure execution, sandbox environments such as Docker containers with strict capabilities can be employed. Alternatively, execution can be restricted to pre-approved pure functions with no side effects or external dependencies.

\textbf{Interoperability gaps}: Frameworks operate in silos, each using incompatible abstractions for agents, tasks, tools, and memory. For example, CrewAI’s task model cannot be directly interpreted by an AutoGen agent, nor can a SmolAgent planner invoke a LangGraph workflow without significant translation. This fragmentation hinders code reuse, tool portability, and seamless system integration. A promising architectural approach is to adopt SOA principles, by wrapping \textit{AI agents as services} to expose their capabilities via RESTful APIs. This enables basic cross-framework interaction, allowing, for example, a LangGraph planner to invoke a CrewAI coder remotely. However, REST lacks the expressiveness for complex agent interaction. To address this, an emerging direction is the use of communication protocols inspired by FIPA-ACL or modern standards like AutoGen’s messaging layer. In future frameworks, combining both RESTful exposure and protocol-level messaging could enable fully interoperable, collaborative agent ecosystems.

\section{Conclusion}

This paper reviews and analyzes major agentic AI frameworks, such as CrewAI, LangGraph, AutoGen, and MetaGPT, focusing on architecture, memory, communication, guardrails, and service computing support. While all aim to support LLM-driven applications, their design priorities vary: some emphasize modularity and memory (e.g., Semantic Kernel), while others focus on collaboration (e.g., AutoGen, ADK) or role-based coordination (e.g., CrewAI). Communication protocols are still evolving, with new paradigms like ACP and Agora suggesting the need for more robust agent-to-agent and agent-to-human dialogue models.

Despite rapid progress, current agentic AI frameworks face several critical limitations that impede their generalizability, composability, and support for service computing. To further advance this field, key directions include establishing standardized benchmarks for objective comparison and reproducibility, as well as developing universal agent communication protocols to enhance interoperability and scalability across frameworks. Another promising direction is incorporating MAS paradigms, such as negotiation, coordination, and self-organization, into existing frameworks.

\bibliographystyle{IEEEtran}
\bibliography{bibilio}
\end{document}